\documentclass[final,12pt]{clear2024} % Include author names

\usepackage{caption}
\usepackage{forest}
\usepackage{graphicx}
\usepackage{svg}
\usepackage{tikz}
\usepackage{transparent}
\usetikzlibrary{arrows.meta}
\usetikzlibrary{positioning,tikzmark}

\newcommand{\pa}{\text{pa}}

\title[Causal discovery in a complex industrial system]{Causal discovery in a complex industrial system: A time series benchmark}
\usepackage{times}
\clearauthor{%
 \Name{S{\o}ren Wengel Mogensen} \Email{soren.wengel\_mogensen@control.lth.se}\\
 \addr Department of Automatic Control, Lund University, Lund, Sweden
 \AND
 \Name{Karin Rathsman} \Email{karin.rathsman@ess.eu}\\
 \addr The European Spallation Source, Lund, Sweden
 \AND
 \Name{Per Nilsson} \Email{per.nilsson2@ess.eu}\\
 \addr The European Spallation Source, Lund, Sweden%
}

\begin{document}

\maketitle

\begin{abstract}%
  Causal discovery outputs a causal structure, represented by a graph, from observed data. For time series data, there is a variety of methods, however, it is difficult to evaluate these on real data as realistic use cases very rarely come with a known causal graph to which output can be compared. In this paper, we present a dataset from an industrial subsystem at the European Spallation Source along with its causal graph which has been constructed from expert knowledge. This provides a testbed for causal discovery from time series observations of complex systems, and we believe this can help inform the development of causal discovery methodology.
\end{abstract}


\section{Introduction}

Datasets from engineered systems are of great value to the causal inference community for several reasons. First, system experts will often know the causal structure, or at least parts of it. Second, operator inputs to the system may be used to emulate interventions. This means that datasets from such systems are useful benchmarks, both for causal discovery and for causal effect estimation. In this paper, we present a benchmark dataset for causal discovery from time series data. This dataset was collected from a complex industrial system at the European Spallation Source, a neutron source facility in Lund, Sweden. It has many properties that makes it interesting for causal discovery. Most importantly, a ground-truth causal graph is known which provides a causal discovery benchmark.

In the remainder of this section, we give a short overview of causal discovery methods (Subsection \ref{ssec:causaldiscovery}) and discuss the use of engineered systems as benchmarks for causal discovery and causal inference (Subsection \ref{ssec:engineeredbenchmarks}). Section \ref{sec:data} describes the data, its collection, the underlying physical system as well as its ground-truth causal graph. Section \ref{sec:analysis} provides visualizations and a simple analysis of the data set. Section \ref{sec:learning} highlights various properties of the dataset and of the underlying system that are important for causal discovery, and Section \ref{sec:discussion} provides a discussion of how the dataset can be used as a causal discovery benchmark. Along with the dataset, we provide R-code to load the data and reproduce the results in this paper.

\subsection{Causal Discovery}
\label{ssec:causaldiscovery}

								\begin{figure}
									%\begin{tabular}{cc}
									\begin{minipage}[.7\textheight]{0.5\linewidth}
										\centering
										\begin{tikzpicture}[scale=0.65]
										\tikzset{vertex/.style
											= 
											{shape=circle,draw,minimum 
											size=1.5em, 
												inner sep = 0pt}}
									%	\tikzset{edge/.style= %{->,> = 
							%			latex',thick}}
										\tikzset{edgebi/.style= {<->,> = 
												latex', 
												thick}}
										\tikzset{every 
											loop/.style={min distance=8mm, 
											looseness=5}}
										\tikzset{vertexFac/.style= 	
											{shape=rectangle,draw,minimum 
											size=1.5em, 
												inner sep = 
												0pt}}
										
										% vertices
										%\draw [line 
										%width=35pt,opacity=0.1,
										% 
										%blue,line 
										%cap=round,rounded
										%corners] (0,0.5) 
										%-- (0,2) -- 
										%(-1.5,1.5) -- 
										%(0,0.5);
										\def\y{-2}						
										\node[vertex] (2) at (-2,2+\y) 	{$2$};
										\node[vertex] (3) at  	(-2,0+\y) {$3$};
										\node[vertex] (4) 	at  (0,0+\y) 	
										{$4$};
										\node[vertex] 	(1) 	at  	
										(0,2+\y)	{$1$};
										
										\node at 
										(-3,1.75+\y) {\Large 
											\textbf{A}};
										
										%edges
										
										\draw[->] (1) to 
										(2);
          \draw[->] (2) to 
										(3);
          \draw[->] (1) to 
										(3);
          \draw[->] (3) to 
										(4);
										
										% graph B
										\def\x{-2}
          \def\z{5}
										\node[vertex] (2) at (-2+\z,2+\x) 	{$2$};
										\node[vertex] (3) at  	(-2+\z,0+\x) {$3$};
										\node[vertex] (4) 	at  (0+\z,0+\x) 	
										{$4$};
										\node[vertex] 	(1) 	at  	
										(0+\z,2+\x)	{$1$};

										%edges
										
										\draw[->] (1) to 
										(2);
          \draw[->] (2) to 
										(3);
          \draw[->] (1) to 
										(3);
          \draw[->] (3) to 
										(4);
          \draw[->, bend left] (4) to 
										(3);
										
										\node at 
										(-3+\z,1.75+\x) {\Large 
											\textbf{B}};

										\end{tikzpicture}
          \captionsetup{format=plain}
          \caption{\textbf{A} is a \emph{directed acyclic graph} (DAG). \textbf{C} is a segment of a time series causal graph. In \textbf{A} and \textbf{C}, each node represents a random variable. In \textbf{B}, each node represents a coordinate process, and \textbf{B} is a \emph{rolled} version of \textbf{C} in which the dependence is summarized at the level of the coordinate processes \citep{danks2013}.}
		\label{fig:exampleGraphs}							\end{minipage}\hspace{.05\linewidth}%
									\begin{minipage}{0.45\linewidth}
										\centering
										\begin{tikzpicture}[scale=0.9]
										\tikzset{vertex/.style = 
											{shape=circle,minimum 
												size=1.5em, 
												inner 
												sep = 0pt}}
										\tikzset{edge/.style = {->,> = latex', 
												thick}}
										\tikzset{edgebi/.style = {<->,> = 
												latex', 
												thick}}
										\tikzset{every loop/.style={min 
												distance=8mm, 
												looseness=5}}
										\tikzset{vertexFac/.style = 
											{shape=rectangle,draw,minimum 
												size=1.5em, 
												inner sep = 0pt}}
										
										% vertices
										%\draw [line width=35pt,opacity=0.1, 
										%blue,line 
										%cap=round,rounded
										%corners] (0,0.5) -- (0,2) -- 
										%(-1.5,1.5) -- 
										%(0,0.5);
										\node[vertex] (11) at  (-4,0) 
										{$X_{t-3}^1$};
										\node[vertex] (12) at  (-2,0) 
										{$X_{t-2}^1$};
										\node[vertex] (13) at  (0,0) 
										{$X_{t-1}^1$};
										\node[vertex] (14) at  (2,0) 
										{$X_{t}^1$};
										\node[vertex] (21) at  (-4,-2) 
										{$X_{t-3}^2$};
										\node[vertex] (22) at  (-2,-2) 
										{$X_{t-2}^2$};
										\node[vertex] (23) at  (0,-2) 
										{$X_{t-1}^2$};
										\node[vertex] (24) at  (2,-2) 
										{$X_{t}^2$};
										\node[vertex] (31) at  (-4,-4) 
										{$X_{t-3}^3$};
										\node[vertex] (32) at  (-2,-4) 
										{$X_{t-2}^3$};
										\node[vertex] (33) at  (0,-4) 
										{$X_{t-1}^3$};
										\node[vertex] (34) at  (2,-4) 
										{$X_{t}^3$};
										\node[vertex] (41) at  (-4,-6) 
										{$X_{t-3}^4$};
										\node[vertex] (42) at  (-2,-6) 
										{$X_{t-2}^4$};
										\node[vertex] (43) at  (0,-6) 
										{$X_{t-1}^4$};
										\node[vertex] (44) at  (2,-6) 
										{$X_{t}^4$};

										\node at (-5,0.25) {\Large 
											\textbf{C}};
										
										%edges
										\draw[->] (11) to 
										(12);
										\draw[->] (12) to  
										(13);
										\draw[->] (13) to 
										(14);
										\draw[->] (21) to 
										(22);
										\draw[->] (22) to  
										(23);
										\draw[->] (23) to 
										(24);
										\draw[->] (31) to 
										(32);
										\draw[->] (32) to  
										(33);
										\draw[->] (33) to 
										(34);
										\draw[->] (41) to 
										(42);
										\draw[->] (42) to  
										(43);
										\draw[->] (43) to 
										(44);

										\draw[->] (11) to 
										(22);
										\draw[->] (12) to  
										(23);
										\draw[->] (13) to 
										(24);
										\draw[->] (11) to 
										(32);
										\draw[->] (12) to  
										(33);
										\draw[->] (13) to 
										(34);
										\draw[->] (21) to 
										(32);
										\draw[->] (22) to  
										(33);
										\draw[->] (23) to 
										(34);
          								\draw[->] (31) to 
										(42);
										\draw[->] (32) to  
										(43);
										\draw[->] (33) to 
										(44);
          								\draw[->] (41) to 
										(32);
										\draw[->] (42) to  
										(33);
										\draw[->] (43) to 
										(34);

										\draw[->] (21) to 
										(33);
										\draw[->] (22) to  
										(34);
										\end{tikzpicture}
									\end{minipage}
									
									%\end{tabular}

								\end{figure}
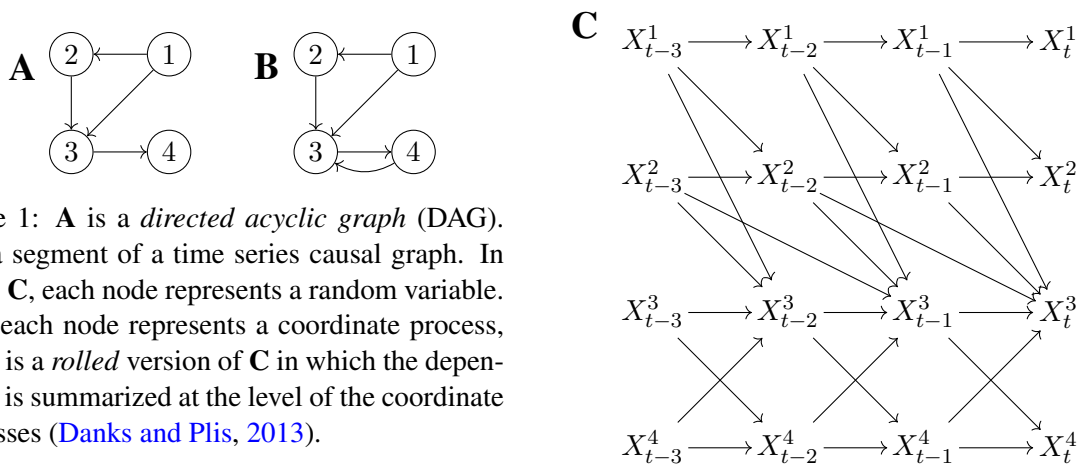

\emph{Causal discovery} methods output causal graphs from observed data. It is common to assume an unknown \emph{directed acyclic graph} such that each node represents a random variable. \cite{spirtes2018search} give an overview of causal discovery in this context. In case of \emph{partial observation}, not every node/random variable is observable. In short, \emph{constraint-based} methods use data to test observable constraints (most commonly conditional independence) that different causal graphs enforce and output graphs based on the observed constraints. \emph{Score-based} methods instead define a certain criterion for the fit between causal structure and observed data, e.g., a penalized version of the maximum likelihood. Score-based algorithms then search for a graph to optimize this criterion. We will now briefly describe one approach to defining a causal model. This will allow us to define what we mean by a causal graph, also in the time series setting.

Assume first that we have a finite collection of random variables, $X_1,X_2,\ldots,X_n$. A \emph{structural causal model} (SCM, \cite{pearl2009, peters2017})  assumes that for each $i = 1,2,\ldots,n$

\begin{align*}
X_i = f_i(X_{\pa_i}, \varepsilon_i)
\end{align*}

\noindent where $\varepsilon_1,\varepsilon_2,\ldots,\varepsilon_n$ are independent random variables and $X_{\pa_i} \subseteq \{X_1,X_2,\ldots,X_n\}$ are the set of \emph{parent variables} of the variable $X_i$. We construct the corresponding \emph{causal graph} on nodes $X_1,X_2,\ldots,X_n$ by including an edge $X_j \rightarrow X_i$ if and only if $X_j \in X_{\pa_i}$. \emph{Acyclicity} is often assumed, i.e., that the causal graph has no directed cycles, $X_{i_1} \rightarrow X_{i_2} \rightarrow \ldots \rightarrow X_{i_{m-1}} \rightarrow X_{i_1}$. An SCM encodes both the observational and interventional distributions of the causal model \citep{pearl2009,peters2017}. It also implies a set of \emph{Markov properties} such that conditional independence can be read off from the causal graph. This is used in many \emph{constraint-based} causal discovery algorithms.

The SCMs can be extended to the time series setting straightforwardly, see \cite{peters2013causal}. We now have a multivariate time series, $X_t = (X_t^1,\ldots,X_t^n)$, and we will assume 

\begin{align*}
X_t^i = f_t^i(X_{\pa_t^i}, \varepsilon_t^i)
\end{align*}

\noindent where $X_{\pa_t^i}$ is the parent set of $X_t^i$.  In this case, the causal dependence, as represented by directed edges, is required to point forward in time, i.e., $X_{\pa_t^j} \subseteq \{X_s^i \}_{s<t,j\in V}$, $V=\{1,2,\ldots,n\}$. One may, of course, impose regularity conditions, e.g., assuming that if $X_{\pa_t^i} = \{X_{t_1}^{i_1}, X_{t_2}^{i_2}, \ldots, X_{t_m}^{i_m} \}$, then $X_{\pa_{t-1}^i} = \{X_{t_1-1}^{i_1}, X_{t_2-1}^{i_2}, \ldots, X_{t_m-1}^{i_m} \}$. We can again define a causal graph, $\mathcal{D}_X$, with nodes $\{X_t^i \}_{t\in \mathbb{Z}, i\in V}$ such that $X_s^i \rightarrow X_t^j$ if and only if $X_s^i \in X_{\pa_t^j}$. This graph is infinite as there is a node for each random variable in the time series. We may use a different graphical representation, a \emph{rolled graph} \citep{danks2013},  such that each node represents a \emph{coordinate process}, $X^i= \{\ldots,X_{-2}^i, X_{-1}^i, X_{0}^i, X_{1}^i, X_{2}^i,\ldots \}$. In this representation $i\rightarrow j$ if and only if there exists $s$ and $t$ such that $X_s^i\rightarrow X_t^j$ in $\mathcal{D}_X$ as defined above. Figure \ref{fig:exampleGraphs}\textbf{C} shows an example of a segment of a causal graph, $\mathcal{D}_X$, when $n = 4$. Figure \ref{fig:exampleGraphs}\textbf{B} shows the corresponding rolled graph (assuming that the edges in the entire $\mathcal{D}_X$ is described by those in the segment shown in the figure). We will think of the ground-truth causal graph (Subsection \ref{ssec:graph}) as a rolled graph of a time series SCM.

\subsubsection{Causal Discovery from Time Series}
When we observe data from a stochastic process, we will most often know the time index of each observation in which case there is a known temporal ordering of the observed values as reflected in the definition of a time series SCM. This is, of course, advantageous in causal discovery as this restrict the possible causal structures significantly. On the other hand, causal discovery from stochastic processes comes with other challenges, some unique to the time series setting. 

In recent years, many methods for causal discovery from time series data have appeared. These methods can be subdivided depending on the assumptions they make. First, there is a natural divide between methods that assume that the underlying causal model evolves in continuous time and methods that assume a causal system evolving in discrete time. Second, methods tend to restrict the model class (e.g., point processes, discrete-time stochastic processes, stochastic differential equations, or (semi)parametric subclasses thereof) they are considering, and the available methods therefore depend on the type of data we wish to analyze. Third, some methods assume \emph{causal sufficiency}, i.e., full observation of the causal system, while others do not. Fourth, methods may assume stationarity or nonstationarity of the observed processes. In the following, we describe some examples of methods with various combinations of these characteristics.

Causal discovery in stochastic processes naturally builds on prior work in the DAG-based causal discovery literature. It is possible to adapt, e.g., the classical FCI-algorithm to the time series setting and further exploit the temporal structure of the time series \citep{chu2008search, entner2010causal, malinsky2018causal}. In stochastic processes, there is also a question of sampling frequency, and \emph{subsampling}, i.e., using a sampling frequency which is lower than the `causal frequency' may pose difficulties for causal discovery which means that specialized methods are required \citep{danks2013,gong2015discovering,Hyttinen2016}. \cite{gong2017causal} consider causal discovery under \emph{temporal aggregation} in which the observed data consists of local averages. Other methods include non-Gaussian structural vector autoregressive models \citep{hyvarinen2010estimation}. It is common to assume stationarity of the observed time series, however, methods that exploit nonstationarity also exist \citep{hyvarinen2016unsupervised}.

Constraint-based algorithms for time series data may either test conditional independence, or use stochastic process-analogues such as \emph{local independence} in continuous-time processes and \emph{Granger causality} in discrete-time processes. \cite{mogensenUAI2018,mogensen2020causal,absar2021discovering} describe causal discovery from continuous-time processes using local independence constraints and \cite{eichler2013} describes causal discovery in discrete-time stochastic processes using Granger causality.

Data sampled from complex dynamical systems is found in many fields of science. In this paper, we consider data from an industrial system. This is mostly due to the fact that such a system may come with substantial expert knowledge of its functioning and therefore provide a useful benchmark. In engineered systems that are less well-understood, causal discovery can augment system understanding. In addition, causal discovery from complex time series has a large potential for other applications, for instance, health registry-based research, Earth system sciences \citep{runge2019inferring}, and economics \citep{hall2022causal}  as these are examples of fields that generate data from high-dimensional, interacting stochastic processes.

\subsection{Engineered Systems as Benchmarks}
\label{ssec:engineeredbenchmarks}

Benchmark data is useful for testing and comparing methods. Good benchmarks should come with a ground truth against which we can compare outputs and they should also resemble the data that we are actually interested in. Obtaining benchmark data with these properties may be difficult. One possibility is the use of \emph{synthetic} data, i.e., using data from computer simulations. \cite{reisach2021beware} show that it may not be easy to simulate realistic data, or that simulated data may show artifacts. This is a concern as causal discovery methods may exploit these artifacts that are products of simulation algorithms and not characteristics of actual causal structures. For this reason, it seems prudent to use real data for causal discovery benchmarking. Of course, in this case the issue is that for real-world data the underlying causal structure may not be known.

We believe that engineered systems may provide useful data in this context. First, their high-level behavior is well-understood as they serve a specific, and known, purpose. Second, experts know how they are constructed which means that the causal structure is fully, or at least partially, known. Third, interventions are common, well-defined, and feasible, for instance, by changing system inputs. In the next section, we describe the dataset that we present. This data comes with an extensive understanding of the underlying system which makes it interesting as a causal discovery benchmark.

\section{Data}
\label{sec:data}

The dataset we provide consists of measurements from a complex system. We explain the context and relevance of this system (Subsection \ref{ssec:ess}). We then describe the system itself in detail (Subsection \ref{ssec:accp}) and the data collection (\ref{ssec:datacollection}). Subsection \ref{ssec:graph} introduces the ground-truth causal graph. Appendix \ref{sec:datastructure} describes metadata and explains how the csv-files containing the data are structured.

\subsection{The European Spallation Source}
\label{ssec:ess}

The European Spallation Source ERIC (ESS) is a neutron source research facility in Lund, Sweden, along with a data center located in Copenhagen, Denmark. The ESS is under construction at the time of writing, however, some systems are already operational. Neutron sources produce beams of neutrons and distribute them to experimental stations. An experimental technique known as \emph{neutron scattering} is then used to study matter through its dispersal of free neutrons, providing a valuable tool for research in, for example, physics, biology, and materials science. Once construction of the ESS is completed, a superconducting linear accelerator will accelerate protons to hit a target which ejects neutrons in a process known as \emph{nuclear spallation}. Upon its completion, scientists will be able to apply for access to the facilities to conduct experiments using various experimental stations. The projected construction cost of the ESS is in the billions of euros, and reliability of the ESS systems is, of course, paramount to minimize system downtime and maximize research output.

\emph{Cryogenics}, that is, cooling to very low temperatures as well as material science at very low temperatures, is essential to the operation of the ESS facility as several components should operate at temperatures close to absolute zero. A \emph{cryoplant} is a system which cools and distributes a coolant. In the following, we describe the accelerator cryoplant \emph{ACCP} which is the largest of three cryoplants at ESS.

\subsection{The ACCP System}
\label{ssec:accp}

The data we provide is sampled from the ACCP system of the ESS which is an industrial refrigeration system. The purpose of the ACCP is to cool the linear accelerator cavities. In a high-level description, coolant (helium) is cooled down to 4.5 Kelvin and distributed to a liquid helium bath that cools the cavities. The coolant is then returned to be recooled, and the overall structure of the system is therefore circular in this sense. In addition, there are several interconnected loops. Figure \ref{fig:diagram} gives a \emph{functional} overview of the system, see Subsection \ref{sssec:diagram} for details on how to interpret this diagram. More details on the ACCP and its construction are provided in \cite{garoby2017european}.

\subsubsection{Functional Diagram}
\label{sssec:diagram}

\begin{figure}
  \centering
  \includegraphics[trim={0 .5cm 0 2.7cm},clip]{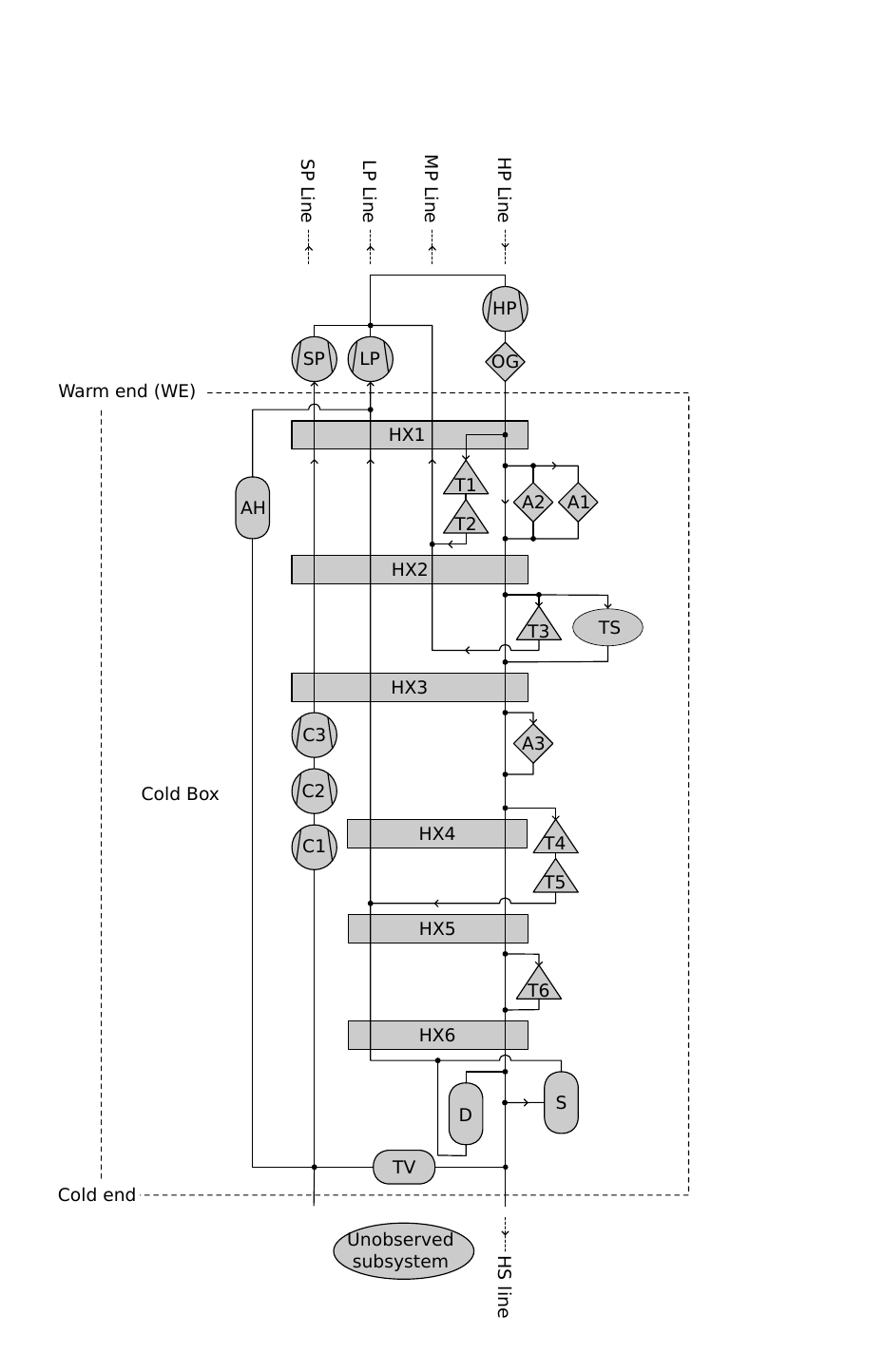}
  \caption{Diagram of the ACCP.}
  \label{fig:diagram}
\end{figure}

Figure \ref{fig:diagram} is provided to illustrate how the ACCP works. In this diagram, each node represents a subsystem and lines indicate a flow of coolant in the direction of the arrow. The section of the ACCP between the two dashed lines in Figure \ref{fig:diagram} is the \emph{cold box}. Starting from the \emph{warm end} (see Figure \ref{fig:diagram}), the ACCP consists of three compressors (SP, LP, and HP), an oil removal and gas management unit (OG), six heat exchangers (HX1, HX2, HX3, HX4, HX5, HX6), six turbines (T1, T2, T3, T4, T5, T6), three adsorbers (A1, A2, A3), a thermal shield (TS), three compressors inside the cold box (C1, C2, C3), a helium subcooler (S), a dewar (D), and a test vessel (TV). From the cold end, the coolant proceeds to a liquid helium bath (HS) where it cools the cavities of the linear accelerator before returning to the ACCP. No measurements are provided from the liquid helium bath, and therefore there is an unobserved subsystem below the ACCP in the diagram. The thermal shield (TS) can also be thought of as partially unobserved as only the pressure, temperature, and flow of the coolant supplied to/from the thermal shield is observed. No measurements from the TS itself are available.

There are three additional subsystems TG, LS and CG that are not shown in the functional diagram. They support more than one subsystem. Turbine General (TG) supports the six turbines, LS the low and sub-atmospheric pressure compressors (LP and SP), and Compressor General (CG) supports the cold compressors (C1, C2 and C3).

A central feature of the ACCP is the circulation of helium. For this reason, it is also natural that some measurements are made between the subsystems in the diagram in Figure \ref{fig:diagram}. The interfaces SPWE, LPWE, MPWE, and HPWE (SP/LP/MP/MP line, warm end) represent measurements made at the warm end dashed line in the diagram, and such measurements are also available in the data. The interfaces will be referred to as subsystems in the remainder, despite the fact that no processes take place there.

While the diagram in Figure \ref{fig:diagram} is useful, it only describes the coolant flow in the ACCP. Subsection \ref{ssec:graph} describes what is believed to be the causal structure of the system. One should note that the subsystems represented are slightly different between the diagram in Figure \ref{fig:diagram} and the causal graph in Figure \ref{fig:causalgraph} (see also Subsection \ref{ssec:graph}). The data set and the remainder of this paper use the subsystems represented in the causal graph in Figure \ref{fig:causalgraph}.

\subsection{Data Collection}
\label{ssec:datacollection}

The ACCP first started operation in the summer of 2020 and has been in intermittent operation since. Operational data has been collected and stored in a central database. The state of each subsystem is described by a number of \emph{process variables} (PVs) that are sampled and recorded repeatedly over time (Table \ref{tab:subsystems}). A PV may, for instance, be a measured temperature, pressure, or flow. We chose three time periods (Period 1: 2022-12-30 17:00 to 2023-01-02 08:00, Period 2:
2023-01-05 19:00 to 2023-01-09 09:00, Period 3:
2023-01-13 16:00 2023-01-16 06:00) and measurements from these three periods constitute the dataset. Over time, the ACCP has been running in different `modes' as operators provide various input values to the system, e.g., set points that are `desired' values for PVs. Periods 1, 2, and 3 were chosen such that these input values are constant within each period, but different across periods. This is related to the notion of \emph{environments} that may represent different interventional settings \citep{peters2017}.

Obviously, a complete description of the system state cannot be collected, however, the PVs that are provided with the dataset are thought to be the most important variables for describing the system state (some subsystem are unobserved, though, see Subsections \ref{sssec:diagram} and \ref{ssec:graph}). The system is expected to evolve quite slowly compared to the sampling frequency.

No data cleaning was done and the dataset is therefore raw data as measured during system operation. We believe that data cleaning is an integral, and nontrivial, aspect of causal discovery, and therefore we present the raw data such that users may experiment with different approaches. One should note that sensor data may be compromised in several ways. Sensors may malfunction or the data acquisition may be noisy for other reasons. Therefore, observed PV values may be outside of the range of possible values. A PV may also be `frozen' at a certain value or make sudden jumps due to sensor malfunctioning.

\begin{table}
    \centering
    \begin{tabular}{c|c|c||c|c|c}
       Index & Description & No. of PVs & Index & Description & No. of PVs \\ \hline
       T1  & Turbine 1 & 5 & C1 & cold Compressor 1 & 5  \\
        T2 & Turbine 2 & 5 & C2 & cold Compressor 2 & 6  \\
        T3 & Turbine 3 & 8 & C3 & cold Compressor 3 & 6  \\
       T4  & Turbine 4 & 5 & CG & Compressor General & 11  \\
       T5  & Turbine 5 & 6 & SP & Sub-atm. Prs. compressor &  17  \\
       T6  & Turbine 6 & 8 & SPWE & SP line Warm End & 2  \\
       TG & Turbine General & 1 & LP & Low Pressure compressor & 16  \\
       A1  & Adsorber 1 80 K & 3 & LPWE  & LP line Warm End & 2  \\
       A2  & Adsorber 2 80 K& 4 & LS  & LP+SP compressor & 8  \\
       A3  & Adsorber 3 20 K& 4 & MPWE  & MP line Warm End &  3  \\
       HX1  & Heat eXchanger 1& 2 & HP & High Pressure compressor & 22   \\
       HX2  & Heat eXchanger 2& 9 & HPWE & HP line Warm End &5 \\
       HX3  & Heat eXchanger 3& 2 &  OG  & Oil and Gas removal &  15  \\
       HX4  & Heat eXchanger 4& 3 &  AH  & Ambient Heater & 5 \\
       HX5  & Heat eXchanger 5& 2 & D  & Dewar &  9  \\
       HX6  & Heat eXchanger 6& 5 & S  & Subcooler& 7 \\
        TV  & Test Vessel & 8 &TS & Thermal Shield$^\ast$ & 8  \\
       HS & Helium Supply$^\ast$ & 6  &   &  &  
    \end{tabular}
    \caption{List of subsystems and number of process variables (PVs) in each subsystem. $^\ast$ indicates that the subsystem is partially unobserved.}
    \label{tab:subsystems}
\end{table}

\subsection{Ground-truth Causal Graph}
\label{ssec:graph}

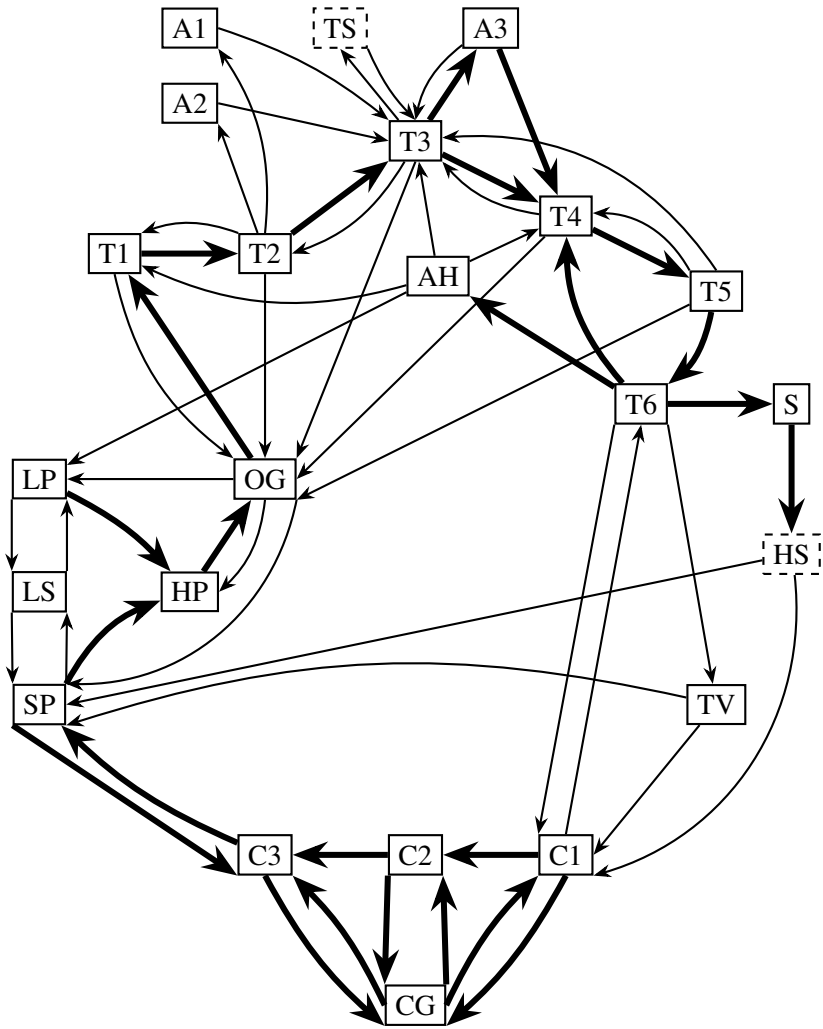
\begin{figure}
    \centering
\begin{tikzpicture}
\newcommand\x{2}
\begin{scope}[every node/.style={rectangle,thick,draw}]
    \node (T1) at (.5*\x,3*\x) {T1};
    \node (T2) at (1.5*\x,3*\x) {T2};
    \node (T3) at (2.5*\x,3.75*\x) {T3};
    \node (T4) at (3.5*\x,3.25*\x) {T4};
    \node (T5) at (4.5*\x,2.75*\x) {T5};
    \node (T6) at (4*\x,2*\x) {T6};
    \node (A1) at (1*\x,4.5*\x) {A1};
    \node (A2) at (1*\x,4*\x) {A2};
    \node (A3) at (3*\x,4.5*\x) {A3};
    \node (B) at (4.5*\x,0*\x) {TV};
    \node (C1) at (3.5*\x,-1*\x) {C1};
    \node (C2) at (2.5*\x,-1*\x) {C2};
    \node (C3) at (1.5*\x,-1*\x) {C3};
    \node (CG) at (2.5*\x,-2*\x) {CG};
    \node (SP) at (0,0) {SP};
    \node (LP) at (0,1.5*\x) {LP};
    \node (LS) at (0,.75*\x) {LS};
    \node (HP) at (\x,.75*\x) {HP};
    \node (OG) at (1.5*\x,1.5*\x) {OG};
    \node (AH) at (2.65*\x,2.85*\x) {AH};
    \node (S) at (5*\x,2*\x) {S};
\end{scope}

\begin{scope}[every node/.style={rectangle,thick,draw,dashed}]
    \node (TS) at (2*\x,4.5*\x) {TS};
    \node (HS) at (5*\x,1*\x) {HS};
\end{scope}

\begin{scope}[>={Stealth[black]},
              every node/.style={fill=white,circle},
              every edge/.style={draw=black,thick}]
    \path [->] (B) edge (C1.east);
    \path [<-] (B) edge (T6.south east);
    \path [->] (T6.south west) edge (C1.north west);
    \path [->] (C1.north) edge (T6.south);

    \path [->] (B) edge [bend left = -15] (SP.south east);

    \path [->] (HS) edge [bend left = 40] (C1.south east);
    \path [->] (HS) edge (SP.east);

    \path [->] (TS.south east) edge [bend right = 10] (T3.north);
    \path [->] (T3) edge [bend right = 0] (TS.south);
    \path [->] (T5.north west) edge [bend right = 30] (T4);
    \path [->] (T5.north) edge [bend right = 30] (T3);
    \path [->] (T5) edge [bend right = 0] (OG.south east);
    \path [->] (T4) edge [bend right = 0] (OG.east);
    \path [->] (T4) edge [bend right = -20] (T3.south east);
    \path [->] (T3.south) edge [bend right = 0] (OG.north east);
    \path [->] (T3) edge [bend right = -20] (T2.east);
    \path [->] (A3) edge [bend right = 20] (T3.north);
    \path [->] (A1.east) edge [bend right = -10] (T3.north west);
    \path [->] (A2.east) edge [bend right = 0] (T3.west);
    \path [->] (T2.north) edge [bend right = 20] (A1.south east);
    \path [->] (T2) edge [bend right = 0] (A2.south east);
    \path [->] (T2.north west) edge [bend right = 20] (T1.north east);
    \path [->] (T1.south) edge [bend right = 20] (OG.north west);
    \path [->] (T2.south) edge [bend right = 0] (OG.north);
    \path [->] (OG) edge [bend right = 0] (LP);
    \path [->] (OG.south) edge [bend right = -20] (HP.east);
    \path [->] (OG.south east) edge [bend right = -40] (SP.north east);
    \path [->] (LP.south west) edge [bend right = 0] (LS.north west);
    \path [->] (LS.north east) edge [bend right = 0] (LP.south east);
    \path [->] (LS.south west) edge [bend right = 0] (SP.north west);
    \path [->] (SP.north east) edge [bend right = 0] (LS.south east);
    \path [->] (AH) edge [bend right = -20] (T1);
    \path [->] (AH) edge [bend right = 0] (T3);
    \path [->] (AH) edge [bend right = 0] (T4);
    \path [->] (AH) edge [bend right = 0] (LP);

\end{scope}
\begin{scope}[>={Stealth[black]},
              every node/.style={fill=white,circle},
              every edge/.style={draw=black, line width=.8mm}]
    \path [->] (T5) edge [bend left = 20] (T6);
    \path [->] (T6) edge [bend left = 20] (T4);
    \path [->] (T4) edge [bend left = 0] (T5);
    \path [->] (A3) edge [bend left = 0] (T4);
    \path [->] (T3) edge [bend left = 0] (T4);
    \path [->] (T3) edge [bend left = 0] (A3);
    \path [->] (T2) edge [bend left = 0] (T3);
    \path [->] (T1) edge [bend left = 0] (T2);

    \path [->] (T6.east) edge [bend left = 0] (S);    
    \path [->] (S) edge [bend left = 0] (HS); 

    \path [->] (T6) edge [bend right = 0] (AH);

    \path [->] (OG) edge [bend left = 0] (T1);

    \path [->] (LP) edge [bend left = 10] (HP);
    \path [->] (HP) edge [bend left = 0] (OG);
    \path [->] (SP.north east) edge [bend left = 20] (HP);
    
    \path [->] (SP.south west) edge [bend left = 0] (C3.south west);
    \path [->] (C3) edge [bend left = 10] (SP);

    \path [->] (C2) edge [bend left = 0] (C3);
    \path [->] (C1) edge [bend left = 0] (C2);
    \path [->] (C3.south) edge [bend left = -10] (CG.south west);
    \path [->] (CG.west) edge [bend left = -10] (C3.south east);
    \path [->] (C2.south west) edge [bend left = 0] (CG.north west);
    \path [->] (CG.north east) edge [bend left = 0] (C2.south east);
    \path [->] (CG.east) edge [bend left = 10] (C1.south west);
    \path [->] (C1.south) edge [bend left = 10] (CG.south east);
\end{scope}
\end{tikzpicture}
    \caption{Causal graph. A dashed rectangle indicates that the subsystem is unobserved. Thick edges indicate a strong connection.}
    \label{fig:causalgraph}
\end{figure}

In this section, we present the \emph{causal graph} of the system (Figure \ref{fig:causalgraph}). The causal graph was constructed by system experts from their knowledge of the system. The causal structure is represented by a graph at the subsystem level, i.e., every node in the causal graph represents a subsystem, i.e., a collection of PVs (in the example Figure \ref{fig:exampleGraphs}\textbf{B}, each node represented a single coordinate process). There is a total of 35 subsystems, each of which is a collection of physical components (see Table \ref{tab:subsystems}). As explained below only a subset of them is used in the causal graph. Two of the subsystems are (partially) unobserved (HS and TS), however, the unobserved systems are quite sparsely connected to the observed systems (Figure \ref{fig:causalgraph}). Subsystems SPWE, LPWE, MPWE, HPWE, and the heat exchangers (HX) are not represented in the causal graph. The HXs are split between the turbine systems, the SPWE is measured along the edge C3 $\rightarrow$ SP, LPWE is measured along between T5 and LP, MPWE is measured between T2/T3 and LS, and HPWE is measured along the edge from OG to T1. The dewar (D) is included in the T6 system based on background knowledge and therefore not shown in the causal graph. The TG system interacts with all the turbines (T1, T2, T3, T4, T5, T6). However, only a quite weak dependence is expected in stable operations and therefore TG is also not shown in the causal diagram.

Each edge in the graph denotes a direct causal link, in most cases an edge also represents a direct physical connection, e.g., the flow of coolant from one subsystem to another. Even though this graph is directed (no bidirected edges), there could in principle be confounding by unmeasured processes. However, the system is self-contained, safe for the sparsely connectected, partially unobserved subsystems as indicated in Figures \ref{fig:diagram} and \ref{fig:causalgraph}, and system experts believe any confounding to be negligible. Each edge in the causal graph is annotated with `edge strength' based on system knowledge (weak or strong link). In Figure \ref{fig:causalgraph}, thick edges correspond to strong causal connections, and thin edges correspond to weak causal connections. An adjacency matrix, $A$, is available along with the dataset. Each row and column of this matrix corresponds to a subsystem. For subsystem $i$ and $j$, $A_{ij} = 0$ indicates that there is no edge from $i$ to $j$, $A_{ij} = 1$ indicates that there is a weak edge from $i$ to $j$, and $A_{ij} = 2$ indicates that there is a strong edge from $i$ to $j$.

As the causal graph was constructed from background knowledge, it is not possible to ensure its `correctness'. It is not even obvious that a single `correct' causal graph can be specified for a problem of this complexity. The division into subsystems could also be done in different ways. However, the causal graph which we present as the ground truth was constructed by engineers that have worked extensively with this system. The system is complex, however, it is also well-documented and as been running intermittently since 2020. Moreover, the design used in the ACCP has been used for more than 20 years. As the system circulates coolant many connections and interdependencies are obvious from the system layout.

\section{Analysis}
\label{sec:analysis}

\begin{figure}
\centering
\includegraphics[scale = .82]{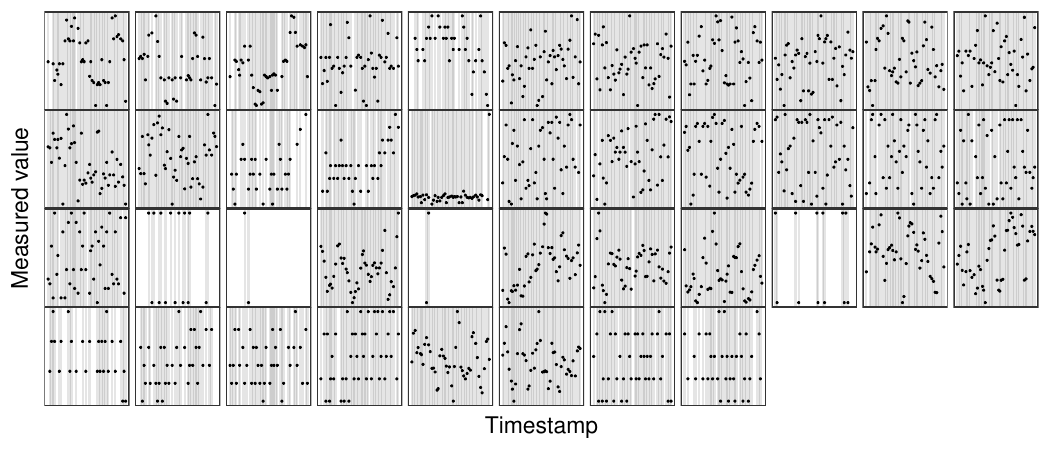}
\caption{Raw data from 10 seconds (41 PVs). The vertical axis is different across subplots. Each point indicates an observation (timestamp and measured value). A vertical line has been added through each point to highlight the irregular nature of the sampling.}
\label{fig:irregular}
\end{figure}

In this section, we visualize the data and compute some simple statistics. The dataset consists of measurements from 233 time series, all of which are measured values, e.g., temperatures, pressures, and flows. We denote an observed value by $x_{t}^{p}$ where $t$ is a time index in $[0,1]$ (after normalization of the time interval) and $p \in \{1,\ldots,233 \}$ is a process index. That is, $x_t^{p}$ is the measured value of process $p$ in subsystem $s$ at time $t$. We use $X^{p}$ to denote a stochastic process corresponding to the observations $x_t^{p}$, and we say that $X^{p}$ is a \emph{process variable} (PV). We let $\mathcal{T}_p$ denote the set of time points, $t$, such that we have observed process $p$ at time $t$. The data is not sampled \emph{regularly}, i.e., $\mathcal{T}_p$, $\mathcal{T}_{p'}$ need not be equal when $p\neq p'$. Moreover, the sampling frequency changes also within each PV, see Figure \ref{fig:irregular}. This figure also highlight the fact that some PVs seem to change their values in a discrete manner, possibly due to the precision of measurement.

\begin{figure}
\centering
\includegraphics[scale = .82]{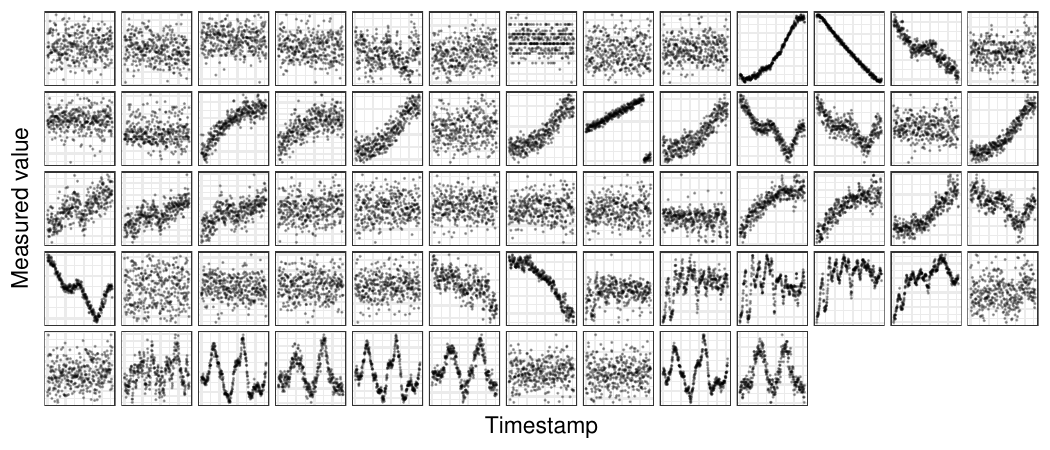}
\caption{Aggregated and subsampled data from 62 PVs during one hour of operation. The vertical scale is different across subplots.}
\label{fig:pvPlot}
\end{figure}

The PVs exhibit different behaviors which is to be expected from the fact that they represent different types of measurements, e.g., temperatures, pressures, and flows (measurement units are provided in the metadata). Figure \ref{fig:pvPlot} shows 62 PVs over the first hour of period 1. We see that some seem to be pure noise, some show a strong trend, and some oscillate (data in the figure is aggregated for each second, and then subsampled to one observation per 10 seconds). One process shows a sudden drop. Some PVs, e.g., the first three in the first row in Figure \ref{fig:pvPlot}, look like noise during the first hour, but show clear trends over the 55-hour span in Figure \ref{fig:pvPlotLongAggr} in Appendix \ref{app:datadescription}.

The sampling frequency is known to be quite high, so it may make sense to reduce the data. It is, for example, possible to subsample the time series (for $p$ consider only observations $\tilde{\mathcal{T}_p} \subseteq \mathcal{T}_p$) or aggregate them over intervals, e.g., compute average PV values for each second of observation to achieve time series with observations every second. In the following correlation and Granger causality analysis, we have done the latter.

Figure \ref{fig:corrplot} in Appendix \ref{app:datadescription} presents a correlation plot (Pearson correlation) of a subset of the PVs in period 1. One should note that strong correlations may be \emph{spurious} as they ignore the autocorrelation. On the other hand, we do observe that some PVs that correspond to the same subsystem are highly correlated which seems reasonable as they describe different aspects of the same system component. Similarly, Figure \ref{fig:grangPlotSec1HourLag50} in Appendix \ref{app:datadescription} shows $p$-values from tests of Granger noncausality. For an ordered pair of subsystems, $(i,j)$, this tests if the past of $j$ is predictive of the present of $i$ when conditioning on the past of $i$. This simple approach also recovers some of the structure from the causal graph.

\section{Learning from ACCP Data}
\label{sec:learning}

We imagine that the ACCP dataset will be useful for benchmarking causal discovery methodology. In this section, we discuss ways to use the data and the challenges that various properties of the data pose for causal discovery.

The most straightforward use of the dataset, and its causal graph, is to compare output graphs from causal discovery with the causal graph. One variation is to assume partial observation such that only some subsystems are observed. \cite{Mogensen2020a} use \emph{directed mixed graphs} as graphical representations of partially observed systems. In this case, such graphs could be the learning target and they can easily be computed from the causal graph given a partition of the subsystems into observed/unobserved. It is also possible to input different types of information to a structure learning algorithm, e.g., the subsystem grouping may be taken as prior knowledge. Alternatively, one can input partial knowledge of the causal graph and use data to refine this knowledge.

The sampling frequency is quite high and this means that subsampling is possible, maybe even advantageous. Comparisons can be made for different subsampling frequencies. Other types of data preprocessing may be important before the application of a causal discovery algorithm, and this may also be tested.

The dataset is divided into three time periods. In each time periods, the ACCP system ran without operators making any changes to input parameters. On the other hand, input parameters are different across the three time periods (the value of these parameters are not provided with the dataset). It may be possible for causal discovery algorithms to leverage this information \citep{peters2017}.

\subsection{Challenges}
The dataset represents a number of challenges that are common when learning causal structure from real-world time series data. In the following, we discuss the properties that may pose issues when analyzing the data.

\paragraph{Partial Observation}
The behavior of the physical ACCP system cannot be captured completely by any dataset of a reasonable size. Instead it must be described by certain well-chosen measurements of temperatures, pressures, etc. In causal modeling, the issue of \emph{partial observation} (i.e., whether the entire system is observed) is central. The measurements from the ACCP do clearly not, in themselves, represent the `entire system'. On the other hand, the measured PVs were chosen by the system manufacturer with the purpose of monitoring and understanding system behavior. Moreover, if we consider the ACCP from the level of subsystems, we do almost have full observation as only a couple of subsystems are unobserved, and these subsystems are only sparsely connected to the observed subsystems.

\paragraph{Hierarchical Structure}
The data has a natural hierarchical structure. There is a large number of PVs and these are divided into 35 subsystems. The number of subsystems is manageable for a human to understand, and each subsystem represents a physical and well-defined component of the system. This means that the causal interpretation on this level is quite appealing. On the other hand, data is only available at the lower PV-level, and this means that the causal structure should be inferred from a large number of underlying processes.

\paragraph{Sampling and modeling frequencies}
The underlying physical processes can be conceptualized as continuous-time stochastic processes, however, the sampling is done in discrete time. The sampling frequency is fairly large, typically 14 Hz, and it may be beneficial to choose a lower frequency for modeling.

\paragraph{Cyclic dependence}
The data-generating mechanism has cycles, and in fact the ACCP is dominated by a large cyclic component which roughly corresponds to the flow of the coolant.

\paragraph{Different timescales} Different parts of the system may operate at different timescales which may complicate causal learning.

\section{Discussion}
\label{sec:discussion}

Benchmark data is important to test causal discovery methodology, and real-world data avoids the pitfalls of synthetic data. In the dataset presented in this paper, the ground-truth causal graph is constructed from expert knowledge of the system, and so is the division of observed processes into subsystems. The measurements describe what is believed to be the most important aspects of the physical systems, however, they are, of course, not exhaustive. For these reasons, one has to keep in mind that other graphs may be equally correct in some sense. This necessarily means that if algorithms fare similarly on this dataset, one should be careful not to overinterpret small differences in the performances. Therefore, the benchmark that this paper describes is perhaps best thought of as an opportunity for researchers to apply their methods to a real and complex dataset in which the gist of the causal structure is known.

% Acknowledgments---Will not appear in anonymized version
\acks{Søren Wengel Mogensen was supported by a DFF-International Postdoctoral
Grant (0164-00023B) from Independent Research Fund Denmark. Søren Wengel Mogensen is a member of the
ELLIIT Strategic Research Area at Lund University.}

\bibliography{references}

\newpage

\appendix

\section{Data Structure and Overview}
\label{sec:datastructure}

Metadata is provided in the
file systemoverview.csv (in the following, \emph{the system overview}). This file contains a line for each PV containing

\begin{itemize}
    \item subsystem index (\texttt{Subsystem.Index}),
    \item subsystem description (\texttt{Subsystem.Description}),
    \item sensor type index (\texttt{Sensor.Index}),
    \item sensor type description (\texttt{Sensor.Description}),
    \item PV name (\texttt{PV.Name}),
    \item description of the PV (\texttt{PV.Description}), and
    \item measurement unit (\texttt{PV.Unit}).
\end{itemize}

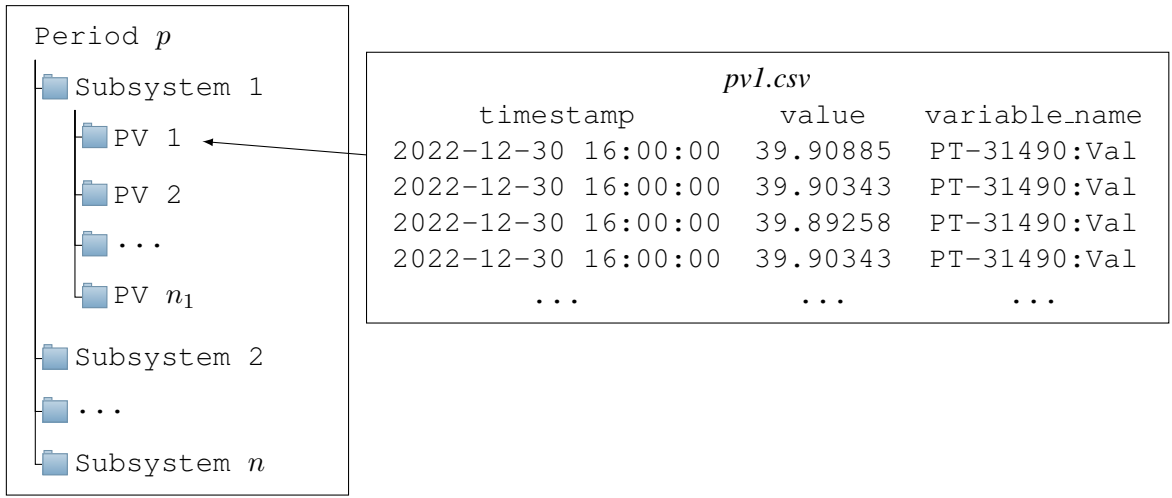
\begin{figure}[ht!]
\centering
    \begin{forest}
      begin draw/.append code={},
      before drawing tree={fit node={tree}{noddies,name=tree}{}},
for tree={
    font=\ttfamily,
    grow'=0,
    child anchor=west,
    parent anchor=south,
    anchor=west,
    calign=first,
    inner xsep=7pt,
    edge path={
      \noexpand\path [draw, \forestoption{edge}]
      (!u.south west) +(7.5pt,0) |- (.child anchor) pic {folder} \forestoption{edge label};
    },
    before typesetting nodes={
      if n=1
        {insert before={[,phantom]}}
        {}
    },
    fit=band,
    before computing xy={l=15pt},
  }
      [Period $p$ [Subsystem 1 [PV 1,name=PV1] [PV 2] [\ldots] [PV $n_1$]] 
      [Subsystem 2 ] [\ldots] [Subsystem $n$ ]]
      \node[noddies] (pv) at (10,-2) {  \begin{tabular}{c c c}
  \multicolumn{3}{c}{\it pv1.csv} \\
   \tt timestamp & \tt value  & \tt variable\_name  \\
  \tt 2022-12-30 16:00:00 & \tt 39.90885 & \tt PT-31490:Val \\
  \tt 2022-12-30 16:00:00 & \tt 39.90343 & \tt PT-31490:Val  \\
 \tt 2022-12-30 16:00:00 & \tt 39.89258 & \tt PT-31490:Val \\
 \tt 2022-12-30 16:00:00 & \tt 39.90343 & \tt PT-31490:Val \\
 \tt \ldots & \tt \ldots & \tt \ldots
  \end{tabular}};
      \draw[-latex] (pv) -- (PV1);
    \end{forest}
  \caption{Overview of data structure. A csv-file provides the data for a single PV in a time period $p$. The data has more decimals than shown.}
  \label{fig:datastructure}
\end{figure}

The dataset is structured in a hierarchy with levels \emph{period}, \emph{subsystem}, and \emph{PV}. The file period\_p/subsystem\_i/pv.csv contains measurements from a single PV from subsystem $i$ in period $p$, see also Figure \ref{fig:datastructure}. Each measurement in such a csv-file is simply a timestamp and a measured value. The units of measurement are different for different PVs as specified in the system overview.

\definecolor{folderbg}{RGB}{124,166,198}
\definecolor{folderborder}{RGB}{110,144,169}
\def\Size{4pt}

\tikzset{
  parent/.style={align=center,text width=4cm,fill=gray!50,rounded corners=2pt},
  child/.style={align=center,text width=2.5cm,fill=gray!20,rounded corners=6pt},
  grandchild/.style={fill=white,text width=2.3cm}
}
\tikzset{
  folder/.pic={
    \filldraw[draw=folderborder,top color=folderbg!50,bottom color=folderbg]
      (-1.05*\Size,0.2\Size+5pt) rectangle ++(.75*\Size,-0.2\Size-5pt);  
    \filldraw[draw=folderborder,top color=folderbg!50,bottom color=folderbg]
      (-1.15*\Size,-\Size) rectangle (1.15*\Size,\Size);
  }
}

\forestset{
  fit node/.style n args=3{
    tempdimxa/.min={x()+min_x()}{#1},
    tempdimxb/.max={x()+max_x()}{#1},
    tempdimya/.min={y()+min_y()}{#1},
    tempdimyb/.max={y()+max_y()}{#1},
    draw tree method/.prefix code/.process=R4w4
      {tempdimxa}{tempdimya}{tempdimxb}{tempdimyb}
      {\node[fit={(##1,##2)(##3,##4)},#2]{#3};},
  },
}
\tikzset{noddies/.style={draw=black,rectangle,fill=white,align=left}}

\newpage

\section{Description of the Data}
\label{app:datadescription}

\begin{figure}[!h]
\centering
\includegraphics[scale = .82]{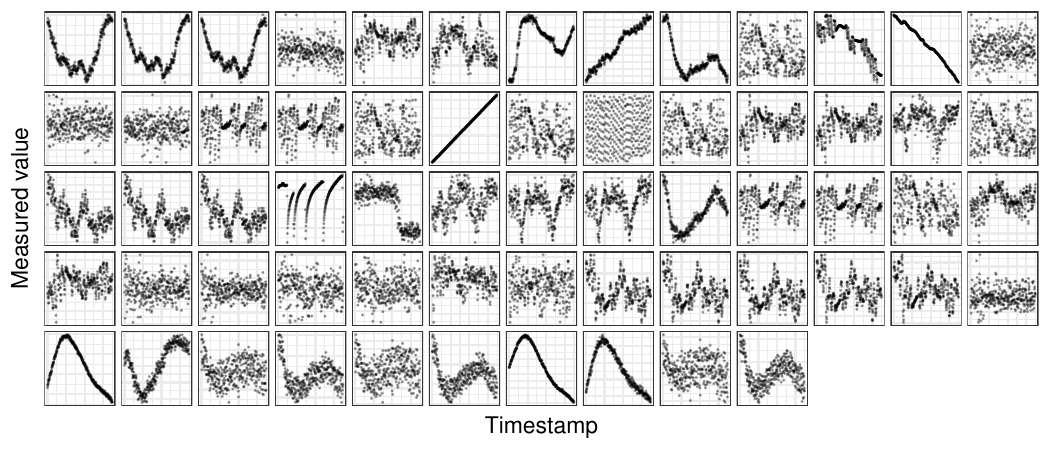}
\caption{Aggregated data (one observation per 10 minutes). The  PVs are the same as in Figure \ref{fig:pvPlot}. This plot covers 55 hours of operation.}
\label{fig:pvPlotLongAggr}
\end{figure}

\begin{figure}
\centering
\includegraphics[scale = .82]{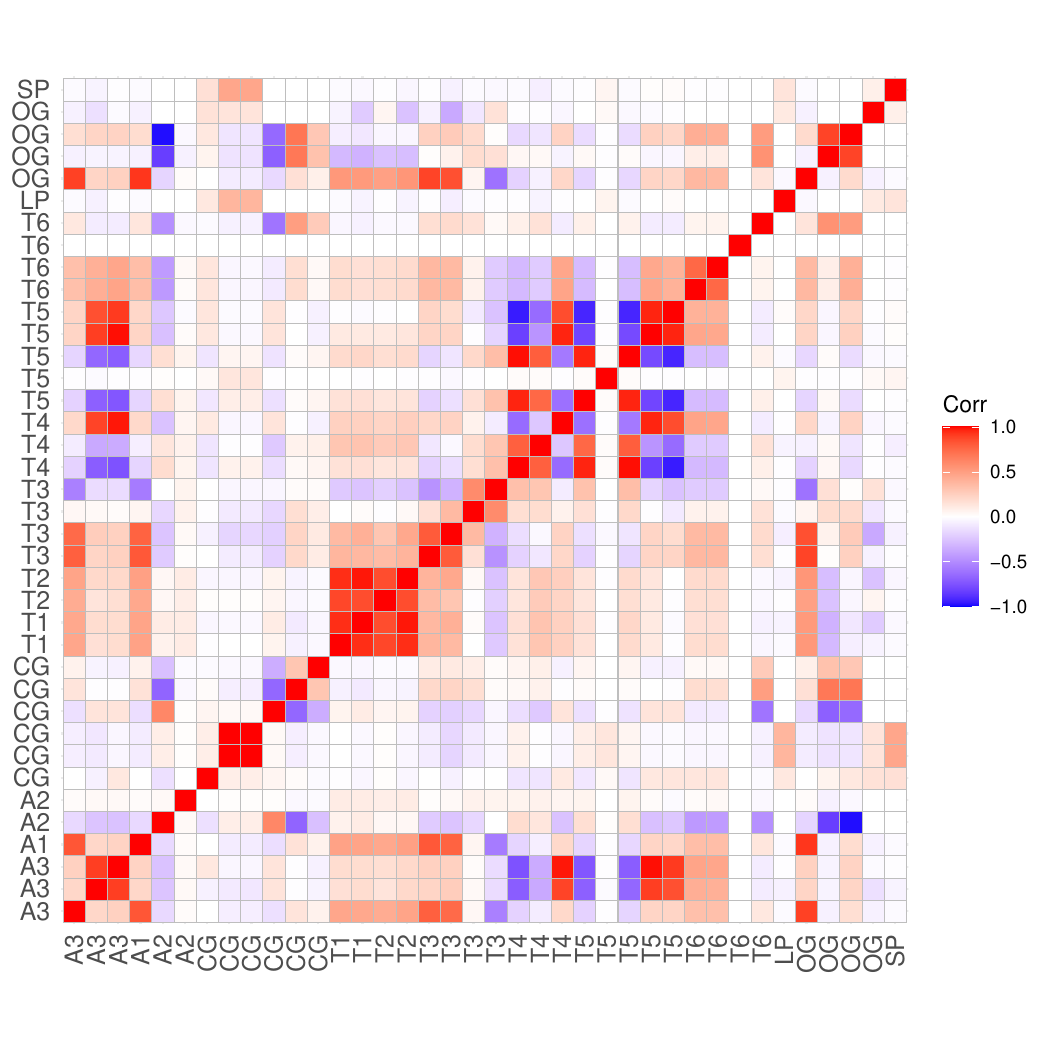}
\caption{Correlation plot of 38 PVs (aggregated data, seconds). At each row and column, the subsystem of the PV is indicated, not the PV-name itself. Pearson correlation was computed for each pair of variables, not accounting for temporal structure. This means that correlations may be \emph{spurious}. As expected from the causal diagram, we see strong correlations between PVs from subsystems A3, T4, and T5, for example.}
\label{fig:corrplot}
\end{figure}

\begin{figure}[!h]
\centering
\includegraphics[scale = .82]{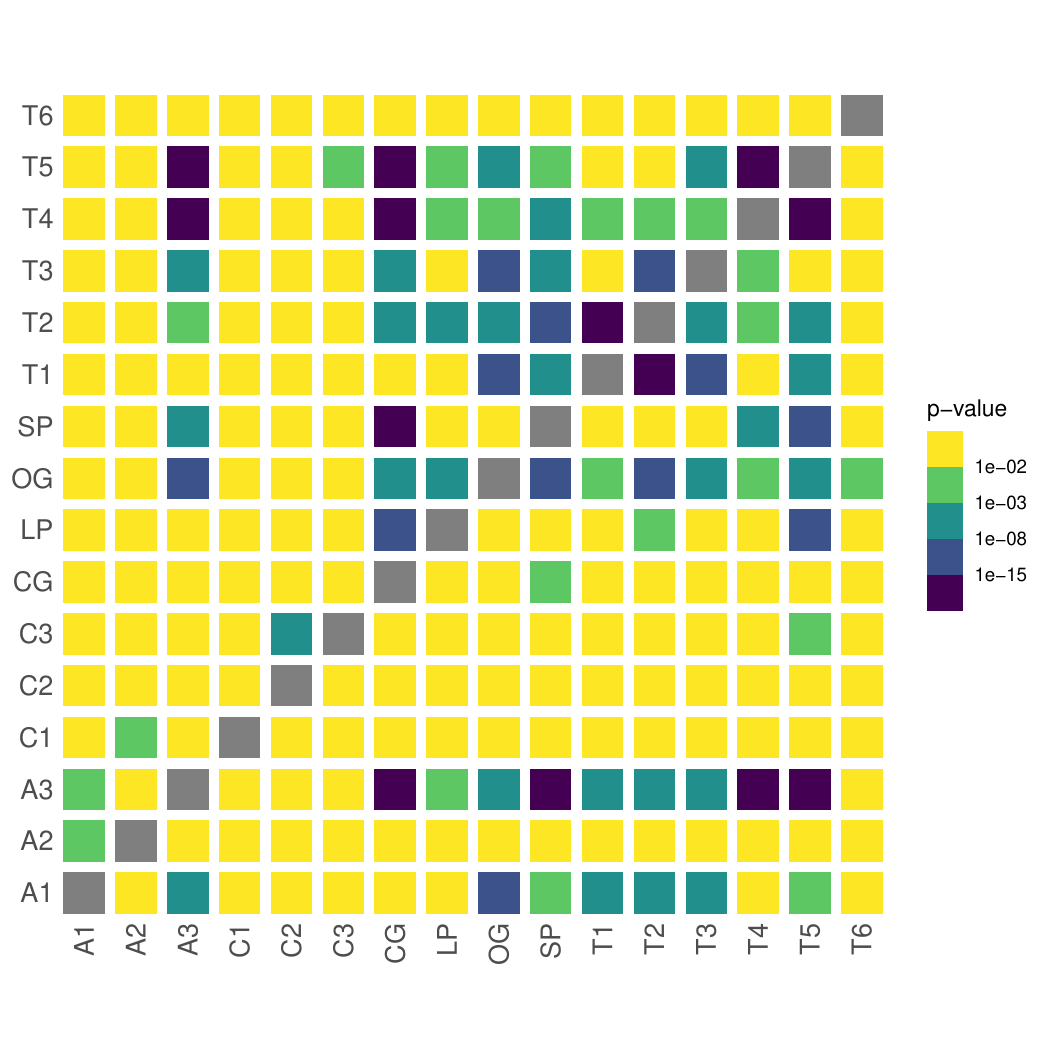}
\caption{Tests of Granger noncausality based on one hour of data (aggregated to seconds). The color scale is discrete: p-values larger than 0.01 are yellow, p-values larger than 0.001 (but less than 0.01) are limegreen, etc. For subsystems $i$ and $j$, the square in $(i,j)$ (row index $i$ and column index $j$) contains the test result for testing if  subsystem $j$ is Granger-noncausal for subsystem $i$.}
\label{fig:grangPlotSec1HourLag50}
\end{figure}

\end{document}